# A novel Neural-ODE model for the state of health estimation of lithium-ion battery using charging curve

Yiming Li[a], Man He[a], Jiapeng Liu[a,*]

[a] Sun Yat-Sen University, School of Advanced Energy, Shenzhen 518107, China
Email: liujp59@mail.sysu.edu.cn

**Abstract**

The state of health (SOH) of lithium-ion batteries (LIBs) is crucial for ensuring the safe and reliable operation of electric vehicles. Nevertheless, the prevailing SOH estimation methods often have limited generalizability. This paper introduces a data-driven approach for estimating the SOH of LIBs, which is designed to improve generalization. We construct a hybrid model named ACLA, which integrates the attention mechanism, convolutional neural network (CNN), and long short-term memory network (LSTM) into the augmented neural ordinary differential equation (ANODE) framework. This model employs normalized charging time corresponding to specific voltages in the constant current charging phase as input and outputs the SOH as well as remaining useful of life. The model is trained on NASA and Oxford datasets and validated on the TJU and HUST datasets. Compared to the benchmark models NODE and ANODE, ACLA exhibits higher accuracy with root mean square errors (RMSE) for SOH estimation as low as 1.01% and 2.24% on the TJU and HUST datasets, respectively.

**Keywords** Neural ordinary differential equation; Machine learning; Lithium-ion batteries; State of health estimation;

# 1 Introduction

Lithium-ion batteries (LIBs), with their high energy density, low self-discharge rate, and long cycle life, have found widespread application in electric vehicles, smart grids, and aerospace systems. However, prolonged cycling causes irreversible side reactions, such as the formation of solid electrolyte interphase (SEI) layer and the growth of lithium dendrites. These phenomena increase internal resistance and accelerate capacity degradation, thereby elevating the risk of system failure [1–3]. To mitigate safety hazards resulting from degradation, accurate prediction of LIB lifespan and health status is critical. The state of health (SOH) is a key metric for estimating the remaining useful life (RUL) of LIBs [4], and 80% of SOH is generally regarded as the threshold of end-of-life (EOL) in vehicles [5].

Existing methods for battery lifetime prediction can be broadly categorized into physics-based and data-driven methods, reflecting different principles and structures. Physics-based methods [6–9] simulate electrochemical processes through numerical solutions of partial differential equations, enabling mechanistic insights into battery aging. However, the practical implementation of physics-based methods remains challenging due to computational costs [10]. Conversely, data-driven approaches do not require a specific physical principle. Machine learning models, trained on historical battery aging data, can directly perform SOH estimation by establishing nonlinear mapping relationships between operational parameters and degradation patterns [11].

Prevalent machine learning methodologies that have been employed in numerous works include support vector machines (SVM) [12], support vector regression (SVR) [13,14], Gaussian process regression (GPR) [15,16], random forests (RF) [17], and neural networks (NN) [18–21]. Among these, neural network-based models have garnered significant research attention. Given the strong correlation between lithium battery degradation and temporal factors, the recurrent neural network (RNN) [22] models, including gated recurrent units (GRU) [23] and long short-term memory network (LSTM) [24,25], have emerged as a popular research direction. In addition, convolutional neural networks (CNN) are also widely employed in SOH estimation, owing to their feature extraction capabilities. Chen *et al.* [26] developed a CNN-LSTM hybrid framework which achieved a maximum prediction accuracy improvement of 58% compared to the native LSTM. In ref [27], researchers further advanced this paradigm by integrating CNN with residual neural networks (ResNet), achieving an $R^2$ score of 0.987. Ma *et al.* [28] implemented a modified whale optimization algorithm for automated hyper-parameter optimization in bi-directional LSTM (Bi-LSTM), achieving a mean square error (MSE) below 0.09, using the battery data from NASA datasets.

Although LSTM and GRU models partially address long-range dependency issues in RNNs, they still exhibit limitations when processing extended sequences or intricate dependencies [29]. Despite the capacity of LSTMs and GRUs to trace historical information, some information loss is inevitable [30]. Conversely, attention mechanisms allow the model to dynamically weight input sequence parts based on their relevance to the prediction task. This can improve the capture of salient features and long-range dependencies. Consequently, model performance is enhanced in complex time-series tasks like SOH degradation [31]. Wei *et al.* [32] developed a graph convolutional LSTM network incorporating a dual attention mechanism, which was adeptly applied to the prediction of both SOH and RUL, achieving high accuracy results with root mean square error (RMSE) of 0.0136 and 5.80 for SOH and RUL on the B6 battery of Oxford dataset, respectively. Yu *et al.* [33] effectively integrated a multi-head attention mechanism with a multi-layer perceptron (MLP), effectively capturing features across diverse time scales during battery aging.

Another significant limitation of previous models is that most studies typically focus on *ad hoc* datasets, lacking generalizability. Pepe *et al.* [34] interpreted the evolution of SOH as a dynamical system where the evolution of SOH is regarded as an ordinary differential equation (ODE). With the help of neural-ODE (NODE) [35] and augmented NODE (ANODE) [36] models, they achieved higher accuracy than the original LSTMs and GRUs in predicting the SOH, demonstrating the promise of NODE in SOH estimation. Inspired by the success, we here propose a new framework based on NODE in combination with an attention mechanism. The proposed model integrates attention mechanisms, CNN, and LSTM within an ANODE architecture to capture complex temporal dynamics. We further compared the model's performance with its counterpart that has not implemented the attention layer and concluded that attention enhances prediction accuracy.

The rest of this paper is organized as follows: Section 2 reviews the process of feature selection, the model architecture utilized in this study is described in Section 3, Section 4 discusses the results of capacity degradation estimation and EOL prediction, and we draw conclusions and further research directions finally.

## 2 Data processing

### 2.1 Datasets

The cyclic aging data used in this study are sourced from datasets provided by four different institutions: Oxford University [37], NASA [38], Tianjin University (TJU) [39], and Huazhong University of Science and Technology (HUST) [40]. Notably, the four datasets exhibit distinct degradation patterns, as illustrated in Fig. 1. This heterogeneity in degradation patterns poses significant challenges for accurate predictions. The Oxford dataset includes eight LIBs that underwent repeated charge and discharge cycles at a constant current of 2C (1.48A). The NASA dataset were tested with 18650-type batteries, where the batteries B0005, B0006, B0007, and B0018 are analyzed here and denoted as A1-A4 to keep consistent with reference [34]. The nine batteries we chose in the TJU dataset comprise 42% $LiNiCoMnO_2$ and 58% $LiNiCoAlO_2$. Finally, the HUST dataset contains 77 batteries that were manufactured by A123 (APR18650M1A). These batteries are labeled by their total cycle numbers, as illustrated in Fig. 1 (d). In total, there are 98 batteries analyzed in this work, and the detailed information of these batteries is displayed in Table 1.

Table 1 The chemical components and basic experiment conditions for four datasets

| Dataset | Cathode materials | Capacity (Ah) | Charging Cut-off voltage (V) | cell # |
|---|---|---|---|---|
| Oxford | LCO/NMC | 0.74 | 4.2 | 8 |
| NASA | LCO/NCA | 2 | 4.2 | 4 |
| TJU | NCM+NCA | 2.5 | 4.2 | 9 |
| HUST | LFP | 1.1 | 3.6 | 77 |

**Note:** The cathode materials and corresponding abbreviations of batteries are as follows: LCO ($LiCoO_2$),

NMC (LiNiMnCoO$_2$), NCA (LiNiCoAlO$_2$), NCM (LiNiCoMnO$_2$), and LFP (LiFePO$_4$).

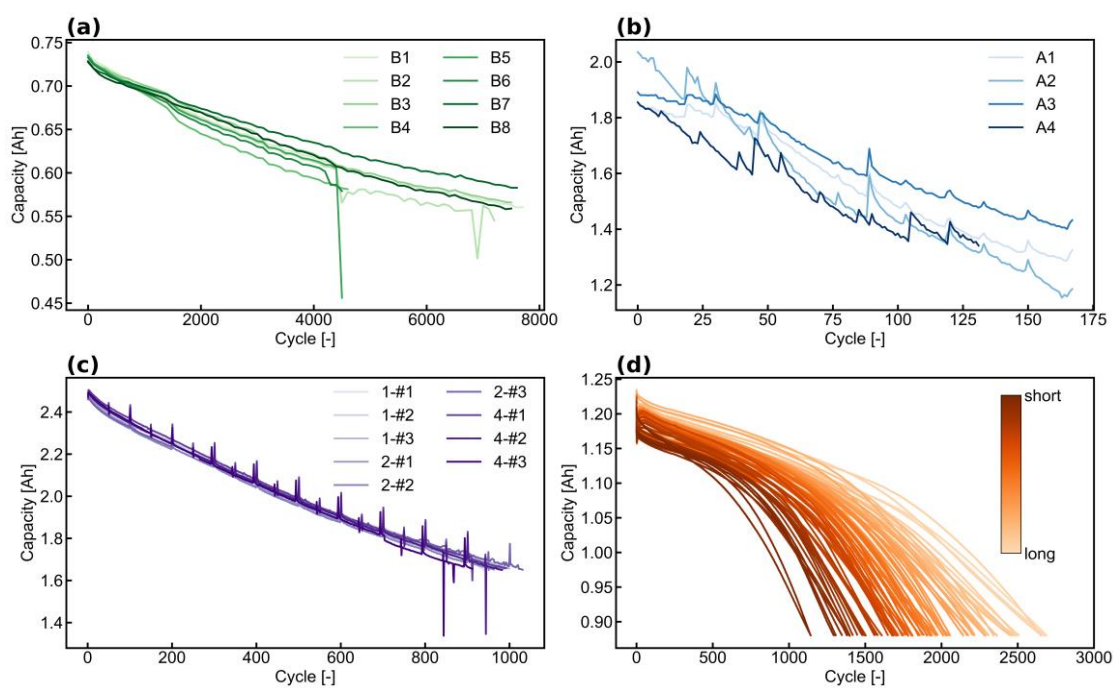

Fig. 1 Capacity curves of all batteries from four datasets: (a) NASA (b) Oxford (c) TJU (d) HUST. The batteries in the HUST dataset are sequenced according to their maximum cycle life, ranging from 2,689 cycles to 1,142 cycles, encompassing a total of 77 batteries.

**2.2 Feature extraction**

Extracting features from the constant current (CC) charging curve is regarded as an effective approach for characterizing the SOH degradation process. It directly reflects aging phenomena such as increased internal resistance and loss of active material, therefore it contains rich health status information. In SOH prediction, the selected features significantly affect final performance. These features must meet specific requirements: (a) they should correlate strongly with the battery's internal degradation state and aging mechanisms, and (b) their extraction should be computationally efficient and robust across different operating conditions and battery types. Based on these standards, the charging profile is suitable because the charging protocol is typically well controlled. This study adopts the method from previous work [28] by extracting the CC process as the model input. Specifically, the input feature includes $SOH_k$, as well as the normalized time corresponding to voltage changes during CC. Firstly, we define the $SOH_k$ as

$$SOH_k = \frac{Q_k}{Q_0} \qquad (1)$$

where $Q_k$ is the capacity at cycle $k$ ($k = 0$ corresponds to the fresh battery).

Regarding the extraction of temporal features, we first divide the charging voltage into $N_V$ equidistant voltages, $(V_1, V_2, \cdots, V_{N_V})$, and then map these voltages into the normalized charging time of $k$-th charging profile. The corresponding discrete time series, $\boldsymbol{t}_k = (t_{k,1}, t_{k,2}, \cdots, t_{k,N_V})^\top$, are then appended to $SOH_k$ to form the input feature vector $\boldsymbol{F}_k$, which has the form of $\boldsymbol{F}_k = (SOH_k, t_{k,1}, t_{k,2}, \cdots, t_{k,N_V})^\top$. Considering that the HUST batteries were charged using two consecutive CC phases, we here adopt a dual-segment sampling strategy with $N_{V,1}$ and $N_{V,2}$ voltages sampled from the 1st and 2nd CC phase, respectively. Detailed sampling data for each dataset are documented in Table 2. Fig. 2 illustrates an example feature extraction for a specified cycle using the Oxford B1 battery and HUST 1-1 battery. To ensure the viability of the feature extraction method, we have also conducted a correlation validation on the battery characteristics, the results of which are illustrated in Fig. 2. As shown in Figure 2(c) and (d), the selected features (normalized time points) show a strong correlation with SOH, validating their effectiveness as inputs for SOH estimation.

Table 2 A comparative analysis of voltage sampling point configurations across various datasets during the CC charging phase.

| Dataset | $N_V$ | Voltage Range (V) | Note |
| --- | --- | --- | --- |
| Oxford | 21 | 3.0 – 4.2 | |
| NASA | 19 | 3.6 – 4.2 | |
| TJU | 19 | 3.6 – 4.2 | |
| HUST | 13 | 3.15 – 3.45 | 1st CC phase |
| | 4 | 3.475 – 3.55 | 2nd CC phase |

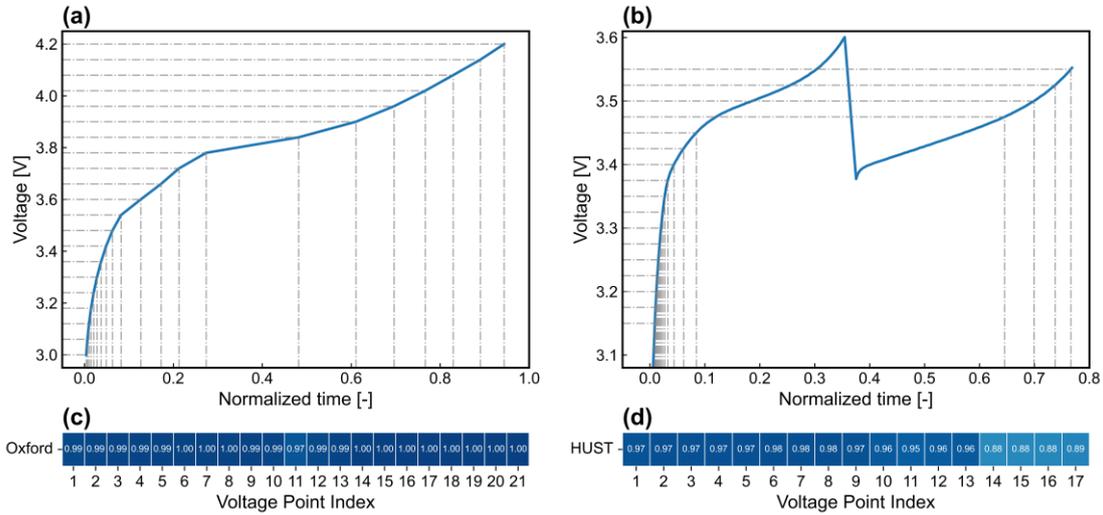

Fig. 2 Illustration of extracted features for the Oxford B1 battery and HUST 1-1 battery: (a) and (b) Voltage versus normalized charging time at the 1000th cycle for the B1 and 1-1 batteries. (c) and (d) The correlation heatmap of various features and their relationship to the SOH within the B1 and 1-1 battery.

The SOH vector, across four datasets, is uniformly defined as $\boldsymbol{SOH} = (SOH_1, \cdots, SOH_k, \cdots, SOH_{N_{tot}})^\top$, where $N_{tot}$ denotes the total number of SOH points for training.

## 3 Methods

### 3.1 Model architecture

As illustrated in Fig. 3, the model contains an attention layer, two CNNs, an LSTM unit, and a linear layer, enriching feature representation of raw data. Subsequently, the ANODE solver is employed for integrating the ODE, which is implemented to model the SOH decay process.

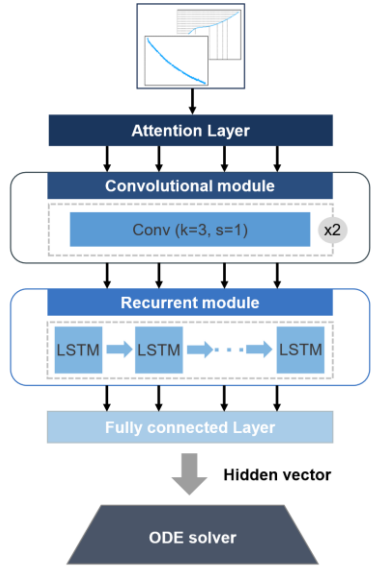

Fig. 3 Schematic diagram of the model architecture

### 3.1.1 Attention mechanism

In the SOH estimation task, the feature vectors extracted from raw data contain rich information at each cycle. However, not all entries contribute equally to accurately predicting SOH. Some entries may better reflect the battery's degradation state than others. To enable an adaptive model to identify and focus on the most informative feature dimensions at each cycle, we implemented the attention mechanism. This mechanism dynamically assigns weights to different features, empowering the model to autonomously learn the significance of different features and thereby concentrating on the most salient ones.

Firstly, the model utilizes a learnable linear transformation to project feature information (derived from the feature vectors, potentially aggregated or represented collectively as $F \in R^{N_{tot} \times (N_V+1)}$) into an $m$-dimensional attention space:

$$A = FW + b \qquad (2)$$

where $W \in R^{(N_V+1) \times m}$ is the trainable weight matrix, $b \in R^m$ denotes the bias vector and $A$ contains the calculated attention scores. This transformation can facilitate dimensionality reduction of the attention representation. More importantly, through the continuous optimization of $W$ during training, it equips the model to learn the importance levels associated with different aspects of the features. The dimension $m$ of this attention space also dictates the number of features that will be modulated in the subsequent step.

Subsequently, the model employs the softmax function to normalize the attention scores row-wise, producing attention weights $\alpha$:

$$\alpha = softmax\ (A, dim = 1) \qquad (3)$$

Here, $\alpha \in R^{N_{tot} \times m}$ has the same dimension with the $A$. This normalization ensures that for each $F_k$, the sum of the $m$ attention weights equals one, forming a probability distribution.

Having calculated the attention weights $\alpha \in R^{N_{tot} \times m}$, the next step is to apply these weights. Here, we propose to impose the attention mechanism on $m$ entries of the original feature vector $F_k \in R^{N_V+1}$. In specific, we choose consecutive $m$ entries from $(t_{k,1}, t_{k,2}, \cdots, t_{k,N_V})^\top$ and denote the sub-feature vector as $F_k^{sub} = (t_{k,p}, t_{k,p+1}, \cdots, t_{k,p+m-1})^\top$, where $1 \leq p \leq N_V + 1 - m$ and $0 \leq m \leq N_V$. Here, $m = 0$ means no attention is imposed on any entry of $(t_{k,1}, t_{k,2}, \cdots, t_{k,N_V})^\top$, whereas $m = N_V$ means attention is uniformly applied across all components of $(t_{k,1}, t_{k,2}, \cdots, t_{k,N_V})^\top$. After implementing the attention score on $F_k^{sub}$ using the element-wise multiplication, we have

$$\boldsymbol{F}_k^{sub} := \boldsymbol{F}_k^{sub} \odot \boldsymbol{\alpha}_{k,:} \tag{4}$$

where $\boldsymbol{\alpha}_{k,:}$ represent the $k$-th row of $\boldsymbol{\alpha}$. We further denote the entries of new $\boldsymbol{F}_k^{sub}$ as $\left(t_{k,p}^{att}, t_{k,p+1}^{att}, \cdots, t_{k,p+m-1}^{att}\right)^\top$. So, the final input feature with attention mechanism imposed on can be formulated as

$$\boldsymbol{F}_k^{att} = \left(SOH_k, t_{k,1}, \cdots, t_{k,p-1}, t_{k,p}^{att}, \cdots, t_{k,p+m-1}^{att}, t_{k,p+m}, \cdots t_{k,N_V}\right)^\top \tag{5}$$

where the superscript "att" indicates that the corresponding entries are imposed with attention mechanisms.

### 3.1.2 CNN-LSTM

CNNs are commonly used for feature extraction in deep learning models. They effectively capture complex relationships between electrochemical processes and battery degradation across cycles [41]. In this work, two 1D convolutional layers are used to extract time-series features.

LSTM units, as an extension of RNNs, address the vanishing gradient problem in RNNs for long-term dependencies. Following the CNN and LSTM layers, a fully connected layer (FC) comprehensively passes the processed data to the ANODE solver.

### 3.1.3 ANODE

Similarly to the NODE framework, ANODE converts neural network forward propagation into a continuous-time initial value problem, as defined in (6). SOH decay is treated as a dynamic evolution system. Thus, the goal of ANODE is to solve the hidden states to predict the SOH.

$$dSOH/d\tau = f(SOH(\tau), \theta_\tau); \quad SOH(0) = SOH_0 \tag{6}$$

here, $f$ is a function of defining the vector field dependent on state $SOH$, and parameters $\theta_\tau$. In addition, ANODE introduces augmented dimensions to address computational inefficiency and training instability in high-dimensional dynamic modeling with traditional NODE. Initially, the augmented dimension is set to zero.

**3.2 Implementation and hyper-parameter setting**

The ANODE model uses an augmented dimension of 20 and applies the default `Dopri5` solver for ODE integration. The adjoint method in ANODE calculates trajectories and gradients independently. The 1D CNN architecture comprises two sequential layers containing 64 and 32 filters, respectively, while the LSTM layer maintains a hidden state of 64 units. The model operates with a batch size of 1. The optimizer combines AdamW and Lookahead [42] for precise model tuning. AdamW serves as the internal optimizer, and Lookahead uses a synchronization period $s = 5$ and the slow weight update rate $\beta = 0.5$, determined via grid search on the validation set. The learning rate starts at 0.01 and follows a three-phase training protocol: (1) a linear learning rate scaling warm-up for 220 iterations, (2) stabilization at the maximum rate for 500 iterations, and (3) a decay phase over 280 iterations, reducing to $1\times10^{-5}$. Fig. 4 illustrates the entire process of this work.

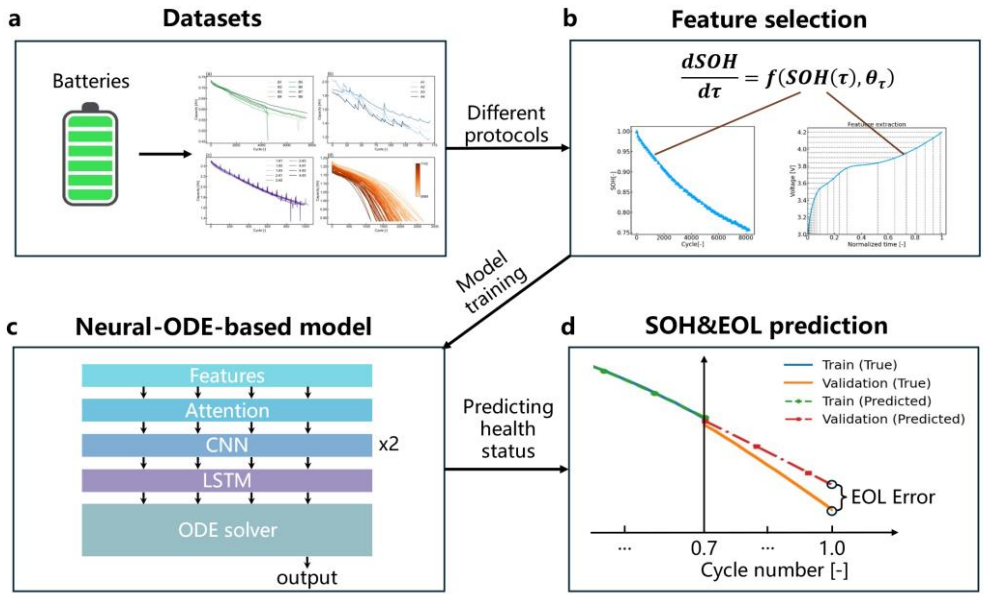

Fig. 4 (a) Various experimental methodologies and diverse chemical compositions result in distinct degradation trends for different batteries. (b) Normalized time feature extraction of voltage from each dataset. (c) Once the entire input set is constructed, it is utilized for model training. (d) Subsequently, the model can accurately predict the performance of batteries across heterogeneous datasets.

## 4 Results

### 4.1 Overview

To facilitate a more equitable comparison with the ANODE and NODE baseline model, we adopted the training approach utilized by Pepe *et al*. [34], selecting a subset of batteries from the NASA and Oxford datasets. For each selected battery, the data were ordered chronologically by cycle number and subsequently divided into a 70% training set and a 30% validation set. The model was trained on the initial 70% of the data, while the validation error derived from the latter 30% were used to refine the hyperparameters. Upon completion of this optimization process, the data corresponding to the remaining batteries within these two datasets were allocated for evaluating the model's predictive performance and for the comparative analysis against the benchmark models.

The loss function $\mathcal{L}_{\mathcal{F}}$ is adapted from the MSE while taking into account the influence of both SOH and charging time. Specifically, it considers the mean error for both the temporal features $t_k$ and the SOH estimate. During the training process, equal significance is attributed to both the SOH and the $t_k$ sequence to ensure a balanced assessment. This helps the model capture dynamic patterns in the input data. The formulation is presented as follows:

$$\mathcal{L}_{\mathcal{F}} = \frac{1}{N_{TP}} \sum_{k=1}^{N_{TP}} \left[ \left( \widehat{SOH}_k - SOH_k \right)^2 + \frac{1}{N_V} |\hat{t}_k - t_k|^2 \right] \quad (7)$$

where $\widehat{(\cdot)}_k$ and $(\cdot)_k$ indicate model prediction and experimental values.

The model's performance is evaluated by calculating the RMSE for the SOH on the validation set, along with the absolute error (AE) for the EOL prediction, as illustrated in (8) and (9)

$$RMSE_{SOH} = \sqrt{\frac{1}{N_{test}} \sum_{k=N_{TP}+1}^{N_{tot}} \left( \widehat{SOH}_k - SOH_k \right)^2} \quad (8)$$

$$AE_{EOL} = \left| \frac{\widehat{EOL} - EOL}{EOL} \right| \quad (9)$$

where $N_{test} = N_{tot} - N_{TP} - 1$ is the number of testing points.

For the validation and generalization capability tests detailed in Section 4, the model's predictions are generated on a per-battery basis. The overall performance metrics are subsequently derived by calculating the error for each individual battery and then determining the mean and standard deviation of these computed errors.

## 4.2 Comparison to native models on the Oxford and NASA datasets

To evaluate our models against established baselines on standard datasets, we adopted the training methodology and specific battery splits from Pepe *et al*. [34] for the Oxford and NASA datasets. Consistent with [34], batteries B1, B3, and B7 from the Oxford dataset, along with A1 and A3 from the NASA dataset were used to model training and parameter tuning. The remaining batteries served as validation samples. Initially, attention was applied to all features in this section. This issue will be further discussed in subsequent sections.

Table 3 Comparison of $RMSE_{SOH}$ (%) prediction errors between the ANODE-improved model and the ANODE and NODE models used in the prior study

| Dataset | Data Split | ACLA Average | Std. | ACL Average | Std. | ANODE Average | Std. | NODE Average | Std. |
|---|---|---|---|---|---|---|---|---|---|
| Oxford | 50 | **1.74** | 0.18 | 2.12 | 0.87 | 2.02 | 1.6 | 2.37 | 0.71 |
|  | 60 | **1.16** | 0.66 | 1.54 | 0.82 | 1.64 | 1.03 | 1.72 | 1.54 |
|  | 70 | 1.74 | 0.94 | 1.89 | 1.21 | 1.77 | 1.52 | **1.13** | 0.71 |
|  | 80 | 1.08 | 0.8 | **1.08** | 0.66 | 1.29 | 0.52 | 1.29 | 0.88 |
|  | 90 | **0.93** | 0.95 | 0.97 | 0.98 | 1.17 | 1.21 | 1.46 | 0.44 |
| NASA | 50 | 8.87 | 0.04 | 7.41 | 5.49 | 10.88 | 11.85 | **4.49** | 0.59 |
|  | 60 | 5.96 | 4.34 | 6.14 | 2.41 | 10.19 | 10.12 | **1.76** | 0.75 |
|  | 70 | **3.54** | 2.39 | 3.96 | 1.65 | 5.36 | 4.17 | 3.97 | 1.07 |
|  | 80 | **2.11** | 1.22 | 2.64 | 0.93 | 3.39 | 2.24 | 2.25 | 0.97 |
|  | 90 | **1.19** | 0.02 | 1.41 | 0.04 | 4.59 | 3.81 | 1.85 | 0.47 |

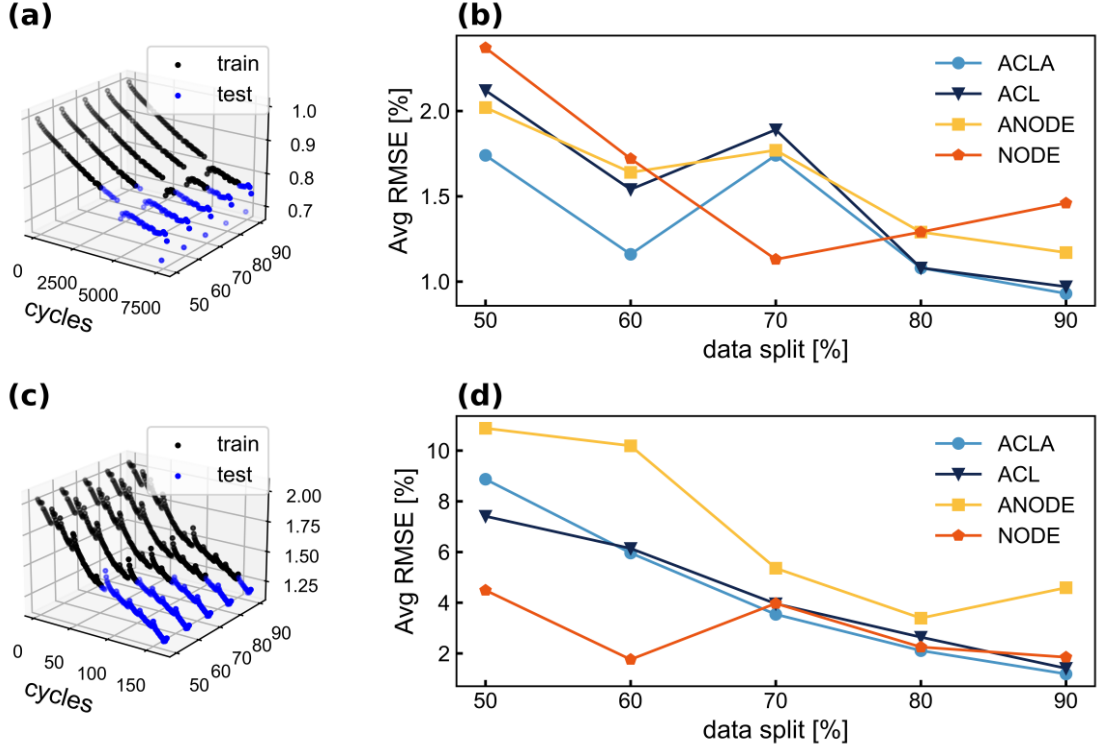

Fig. 5 RMSE of SOH estimation results: (a) and (c) Examples of training and testing points on the capacity curve of the B2 and A2 battery. (b) and (d) Evolution of the test error on the SOH prediction, averaged on the batteries B2, B4–6, and B8 and batteries A2 and A4 with different portions of data used for training (50, 60, 70, 80, 90%).

Table 3 presents the average $RMSE_{SOH}$ and corresponding standard deviations for ACLA, ACL, ANODE, and NODE across varying proportions of training data (50% to 90%). Fig. 5 provides a visual representation of these performance trends. A consistent observation across both datasets is the superior performance of the proposed ACL and ACLA models compared to the original ANODE and NODE. The ANODE model exhibits the poorest performance across multiple test conditions, recording an $RMSE_{SOH}$ of 10.88 at a 50% training data proportion. In contrast, the ACL model exhibits a substantial reduction in error. For instance, at a 90% training proportion on the NASA dataset, its $RMSE_{SOH}$ is 1.41, representing a 69.3% reduction compared to ANODE's 4.59. The ACLA model achieves further optimization, attaining the lowest $RMSE_{SOH}$ of 1.19 under the same conditions. This reflects reductions of approximately 74.1% relative to ANODE and 15.6% relative to ACL, with ACLA consistently surpassing ACL in performance.

The NODE model excels in specific cases—such as an $RMSE_{SOH}$ of 1.13 for 70% Oxford dataset training data, and 4.49 and 1.76 for 50% and 60% NASA dataset training data, respectively. However, the modified ANODE architecture, particularly the ACLA, demonstrates high overall robustness and competitiveness, especially with larger training datasets. For example, with an 80% training proportion on the NASA dataset, ACLA achieves an $RMSE_{SOH}$ of 2.11, outperforming ACL (2.64), ANODE (3.39), and NODE (2.25) by approximately 20.1%, 37.8%, and 6.2%, respectively. These results highlight ACLA's precision in data-rich scenarios.

**4.3 Predictive performance of the model on HUST and TJU datasets**

To further validate the generalization capability of the proposed models, the model were directly used to do estimation on the HUST and TJU datasets, which are entirely independent of the previously utilized Oxford and NASA datasets. Given the substantial volume of data inherent in these new datasets, a uniform sampling approach, guided by cycle count, was implemented to select 80 SOH points from each, serving as inputs for the model evaluation.

**4.3.1 Optimization of attention layers implementation**

Previous experimental investigations demonstrated that applying the attention mechanism comprehensively across all feature points resulted in high prediction accuracy. However, this approach imposed a considerable computational burden. To navigate this accuracy-efficiency trade-off, we explored a more judicious application of attention.

The characteristic charging voltage profile during the CC phase—defined by an initial rapid ascent, a subsequent period of gradual increase, and a final acceleration—contains significant indicators of battery degradation. The recognition that distinct physical stages within this profile, as implied by their aging signatures, possess differential predictive value prompted the development of targeted attention strategies. This framework selectively applies attention mechanisms to key temporal segments rather than uniformly across all feature points, thereby optimizing the extraction of stage-specific information.

We hypothesized that concentrating attention on these specific segments could mitigate the computational load without a substantial sacrifice in predictive performance. Consequently, we re-evaluated the strategy of attention placement and systematically assessed its effectiveness when applied to different intervals along the voltage curve. This refined strategy involved selecting three consecutive feature points from distinct regions representative of the curve's phases, with the aim of capturing key variations indicative of battery health.

Table 4 Comparison of results (%) by different attention layers on HUST and TJU datasets

|  |  | Att_start | Att_mid | Att-end | Att_all |
|---|---|---|---|---|---|
| HUST | $RMSE_{SOH}$ | **2.24** | 2.32 | 2.41 | 2.43 |
|  | $AE_{EOL}$ | **5.30** | 5.57 | 5.78 | 5.83 |
|  | Training time (s) |  | 168 |  | 199 |
| TJU | $RMSE_{SOH}$ | 1.04 | 1.07 | **0.97** | 1.15 |
|  | $AE_{EOL}$ | 1.01 | 1.45 | **0.95** | 1.41 |
|  | Training time (s) |  | 111 |  | 124 |

**Note:** In the Table 4, 'Att_start', 'Att_mid', and 'Att-end' represent the data results when attention layers are implemented at the beginning, middle, and end three points of the feature set, respectively. In contrast, 'Att_all' represents results when attention layers are deployed across the entire feature set.

On the HUST dataset, when attention layers focus on the first three features, the model achieves optimal performance. As shown in Table 4, the $RMSE_{SOH}$ is 2.24 and $AE_{EOL}$ is 5.30. This reflects an accuracy improvement of up to 18% compared to the full-feature application approach. The TJU dataset, however, exhibits the best performance while imposing the attention mechanism on the last three entries of $t_k$, confirming that degradation patterns in different battery systems possess varying feature-sensitive regions. Following a thorough assessment of predictive accuracy and computational efficiency, as documented in the comparative analysis, the Att_start configuration, i.e., imposing the attention mechanism on the first three enties of $t_k$, was selected as the optimal approach for subsequent performance evaluations. This choice also well balances precision and computational cost. Consequently, we adopt this Att_start configuration as the selected optimization strategy for our proposed ACLA model, employing it in all subsequent comparative experiments against baseline methods and state-of-the-art techniques.

**4.3.2 Further Performance Validation of model**

The refined attention configuration strategy was applied to the subsequent validation process. Fig. 6 and Table 5 present a comparison of SOH estimation and EOL prediction results for different models on HUST and TJU datasets. The results clearly demonstrate that the ACLA and ACL models outperform ANODE and NODE in overall prediction performance. On the more challenging HUST dataset, scatter plots visually confirm a strong alignment between the predicted and actual SOH values for ACLA and ACL, whereas NODE noticeably underestimates SOH. Fig. 6(a-h) also demonstrated that the distributions of $RMSE_{SOH}$ of ACLA and ACL model are lower and more centralized. According to Table 5, ACLA demonstrates superior performance on the HUST dataset with the lowest average $RMSE_{SOH}$ of 2.24 and $AE_{EOL}$ of 5.33. These results represent significant improvements over NODE, reducing errors by approximately 57% and 54.7% respectively. Compared to ANODE, which records an $RMSE_{SOH}$ of 3.45 and $AE_{EOL}$ of 8.05, ACLA achieves reductions of 35% and 34% respectively. Furthermore, both ACLA and ACL exhibit notably smaller standard deviations, confirming their enhanced precision and stability.

For the less complex TJU dataset, all models perform well with minimal errors. While NODE achieves a marginally lower $RMSE_{SOH}$ of 0.94, ACLA excels in the critical $AE_{EOL}$ metric with a value of 1.01, outperforming all competing models. These findings underscore the value added by the progressive enhancements to the ANODE framework, particularly the integration of CNN-LSTM-attention modules, leading to substantial performance gains across diverse datasets and metrics.

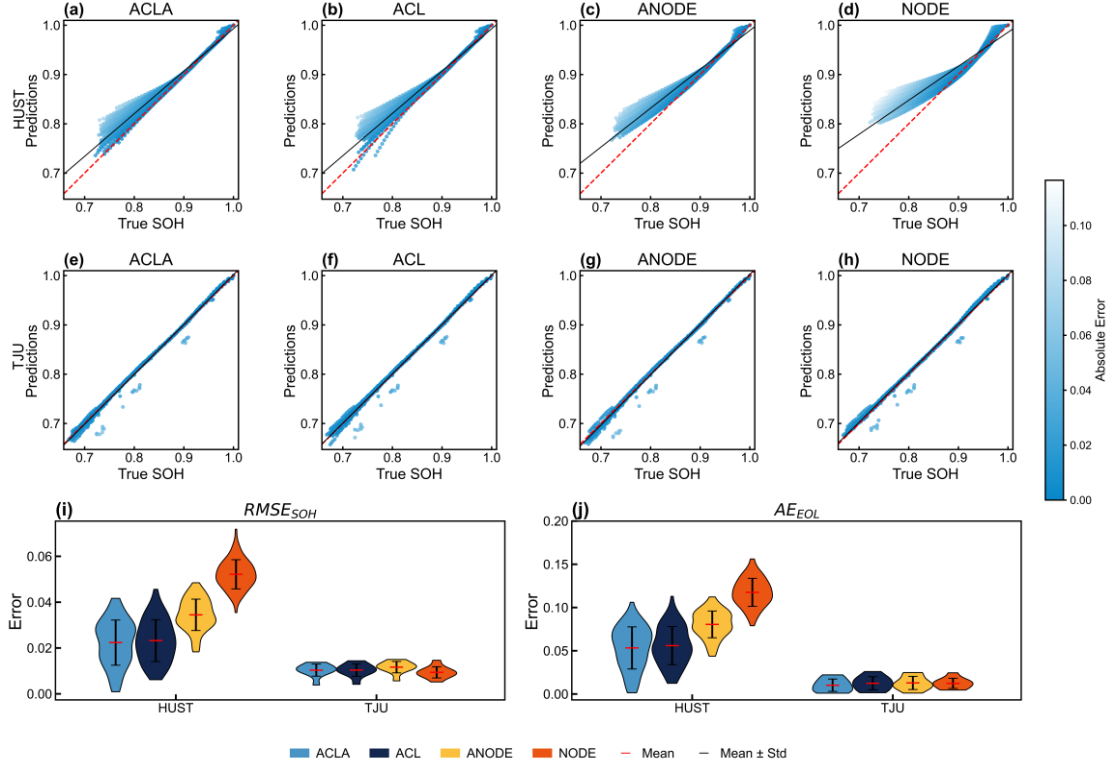

Fig. 6 The illustrations of SOH estimation results in TJU and HUST datasets with different methods.

Table 5 Prediction error (%) of four methods (ACLA, ACL, ANODE, NODE) in TJU and HUST datasets

| Dataset | | ACLA | | ACL | | ANODE | | NODE | |
| --- | --- | --- | --- | --- | --- | --- | --- | --- | --- |
| | | Average | Std. | Average | Std. | Average | Std. | Average | Std. |
| HUST | $RMSE_{SOH}$ | **2.24** | 0.99 | 2.34 | 0.91 | 3.45 | 0.68 | 5.22 | 0.64 |
| | $AE_{EOL}$ | **5.33** | 2.45 | 5.6 | 2.2 | 8.05 | 1.55 | 11.76 | 1.62 |
| TJU | $RMSE_{SOH}$ | 1.04 | 0.27 | 1.04 | 0.27 | 1.17 | 0.25 | **0.94** | 0.24 |
| | $AE_{EOL}$ | **1.01** | 0.72 | 1.24 | 0.76 | 1.28 | 0.76 | 1.23 | 0.59 |

**4.4 Comparison of model errors under different dataset partitions**

To assess the effect of varying training partitions on the prediction error of the proposed ACLA model, its performance was evaluated by adjusting the training data proportion from 50% to 90%. From the Table 6, we can find that even with only 50% of data, the average $RMSE_{SOH}$ on the HUST dataset remains smaller than 6. On TJU dataset, it is lower than 3, and the $AE_{EOL}$ errors are less than 10 in most cases. The results show that ACLA maintains accurate SOH and EOL predictions even with limited training data. This capability is valuable for applying NODE-based models, because early-stage predictions are typically more challenging. This underscores the model's efficiency and suggests the effectiveness of its architecture data-scarce scenarios.

Table 6 Prediction results (%) upon varying methodologies of training set partitioning

| Dataset | Data Split | ACLA | | ACL | | ANODE | | NODE | |
|---|---|---|---|---|---|---|---|---|---|
| | | $RMSE_{SOH}$ | $AE_{EOL}$ | $RMSE_{SOH}$ | $AE_{EOL}$ | $RMSE_{SOH}$ | $AE_{EOL}$ | $RMSE_{SOH}$ | $AE_{EOL}$ |
| HUST | 50 | 5.55 | 14.00 | 6.02 | 15.62 | 5.02 | 13.9 | 7.96 | 18.96 |
| | 60 | 3.06 | 8.00 | 3.47 | 9.52 | 4.35 | 11.17 | 6.38 | 15.89 |
| | 70 | 2.24 | 5.30 | 2.34 | 5.60 | 3.45 | 8.05 | 5.22 | 14.13 |
| | 80 | 1.56 | 3.33 | 1.57 | 3.38 | 2.84 | 5.87 | 4.45 | 8.94 |
| | 90 | 0.78 | 1.39 | 0.79 | 1.45 | 1.88 | 3.31 | 3.42 | 5.76 |
| TJU | 50 | 2.77 | 7.10 | 2.79 | 7.32 | 2.65 | 6.12 | 1.03 | 1.22 |
| | 60 | 1.87 | 3.55 | 1.94 | 3.76 | 1.99 | 3.83 | 1.02 | 0.96 |
| | 70 | 1.04 | 1.01 | 1.04 | 1.65 | 1.17 | 1.28 | 0.94 | 1.23 |
| | 80 | 1.26 | 1.05 | 1.36 | 1.28 | 1.29 | 1.23 | 1.36 | 1.31 |
| | 90 | 0.45 | 1.34 | 1.36 | 0.42 | 1.39 | 0.43 | 1.34 | 0.62 |

## 4.5 Comparative Analysis with Existing Literature

To further demonstrate the performance of ACLA, we compared its $RMSE_{SOH}$ to that of a recently published model based on physical informed neural network [5]. To ensure a reasonable comparison, the data partitioning methodology was strictly aligned with that of ref [5], utilizing a 60% training set, a 20% validation set, and a 20% test set and we compare their results for small sample. The resulting error metrics from this adjusted partitioning approach are reported in Table 7.

Table 7 Comparison of test $RMSE_{SOH}$ (%) with ref [5] on HUST and TJU datasets

| Dataset | Results of ref [5] | | Our results |
|---|---|---|---|
| | 1# | 2# | |
| HUST | 4.85 | 2.02 | 4.04 |
| TJU | 1.21 | 2.02 | 2.44 |

**Note:** In the table, "1#" and "2#" denote the outcomes from ref [5] utilizing one or two batteries, respectively, as training samples.

On the challenging HUST dataset, when trained using data from only a single battery, our ACLA model achieves an $RMSE_{SOH}$ of 4.04. Notably, this result is lower than the reported single-battery performance of 4.85 for the model in ref [5], highlighting ACLA's effectiveness even in data-limited scenarios against an established technique. On the TJU dataset, the reference model [5] achieved a lower error of 1.21 with a single training battery compared to ACLA's 2.44. This suggests that for datasets with potentially simpler degradation dynamics, their approach might be particularly advantageous in the single-battery scenario.

Overall, this comparative analysis demonstrates that ACLA is competitive with state-of-the-art methods in small-sample cases. Its particularly strong performance on the demanding HUST dataset under the single-battery training condition underscores the robustness and potential of our proposed architecture.

## 5 Conclusion

This study proposed and validated ACLA, an innovative framework integrating ANODE with CNN, LSTM, and attention mechanism, for lithium-ion battery SOH assessment. By leveraging the strengths of these components and using discretized charging time features, ACLA provides a robust and accurate method for SOH prediction.

A key contribution is the demonstration of significantly improved generalizability. Extensive validation was conducted using battery datasets from four distinct institutions.

The proposed ACLA model consistently achieved competitive prediction performance on unseen data, effectively addressing a common limitation of existing SOH estimation methods. For instance, validation on the challenging HUST dataset yielded a low average $RMSE_{SOH}$ of 2.24%, showcasing the model's ability to adapt to diverse battery chemistries and degradation patterns. ACLA consistently outperformed baseline models (NODE, ANODE) and its variant without attention (ACL), highlighting the synergistic benefits of its components, especially achieving substantial error reductions on demanding datasets. The model also proved robust, maintaining high accuracy (e.g., $RMSE_{SOH}$ below 3.1% on HUST) even with significantly reduced training data, and showed competitiveness against state-of-the-art small-sample methods. In summary, ACLA offers an effective deep learning architecture for SOH estimation, featuring high accuracy, strong generalization, and robustness.